\documentclass[10pt,twocolumn,letterpaper]{article}
\usepackage[pagenumbers]{cvpr}
\usepackage{url}
\usepackage{microtype}
\definecolor{cvprblue}{rgb}{0.21,0.49,0.74}
\usepackage[pagebackref,breaklinks,colorlinks,allcolors=cvprblue]{hyperref}
\graphicspath{{figures/}}
\usepackage{colortbl}

\definecolor{apmrow}{HTML}{EDF5F4}
\definecolor{resultgain}{HTML}{19764C}
\definecolor{resultdrop}{HTML}{B4483D}

\def\confName{CVPR}
\def\confYear{2027}

\title{Distilling World Action Models for Efficient Bimanual Manipulation \\ via Anchored Path Matching}
\author{
Jialong Liu$^{1,2}$\quad Junjie He$^{2}$\quad Yandu Chen$^{1,3}$\quad Bin Zhou$^{1,4}$\\
Heng Su$^{1,5}$\quad Peijun Tang$^{1}$\quad Jianan Wang$^{1}$\quad Changhao Chen$^{2,\mbox{\includegraphics[height=0.55em]{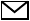}}}$\\[4pt]
{\small $^{1}$Astribot\quad $^{3}$Harbin Institute of Technology\quad $^{5}$Chongqing University}\\
{\small $^{2}$The Hong Kong University of Science and Technology (Guangzhou)}\\
{\small $^{4}$The Hong Kong University of Science and Technology}\\[4pt]
{\small \includegraphics[height=0.7em]{icons/envelope.pdf}\enspace Corresponding author: \href{mailto:changhaochen@hkust-gz.edu.cn}{changhaochen@hkust-gz.edu.cn}}\\[4pt]
{\small \raisebox{-0.15em}{\includegraphics[height=1.1em]{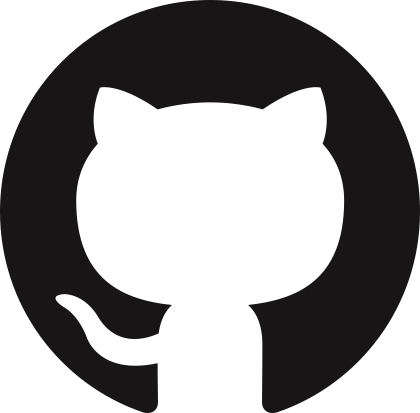}}\enspace Code: \url{https://github.com/JIAlonglong/Anchored-Path-Matching}}\\
{\small \raisebox{-0.15em}{\includegraphics[height=1.1em]{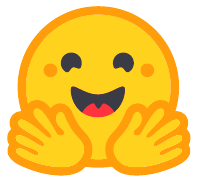}}\enspace Hugging Face: \url{https://huggingface.co/JIAlonglong/Anchored-Path-Matching}}
}
\begin{document}
\raggedbottom
\maketitle
\begin{abstract}
World Action Models (WAMs) jointly generate future video and robot actions through iterative denoising. Few-step distillation reduces the number of model evaluations, but its coarse updates can move the student away from the teacher's trajectory. We propose Anchored Path Matching (APM), which supervises endpoint and short-transition predictions from states reached by the student. From each state, we continue the teacher to zero noise and use its endpoint and first intermediate state as targets. Modality-routed adapters learn to match both targets from the same starting state. When the two gradients conflict, anchor-priority updates preserve the endpoint gradient and remove the opposing component of the local gradient within each modality. One checkpoint supports 1-, 2-, and 4-step joint generation. The sampling budget can be chosen at deployment without retraining or teacher queries. On RoboTwin 2.0, we train only on Clean demonstrations and evaluate directly on the domain-randomized Random setting. At one step, APM improves closed-loop success over continued base distillation at matched inference budgets by 2.0 percentage points on Clean and 1.4 on Random. Video peak signal-to-noise ratio (PSNR) rises from 30.169 to 32.317 dB, with gains in structural similarity (SSIM) and learned perceptual image patch similarity (LPIPS). On two real-robot tasks, APM improves success over the one-step teacher and reduces control-cycle latency by $21.3\times$ relative to the eight-step teacher.

Project page: \url{https://jialonglong.github.io/Anchored-Path-Matching/}.
\end{abstract}
\begin{figure}[t]
\centering
\includegraphics[width=\columnwidth]{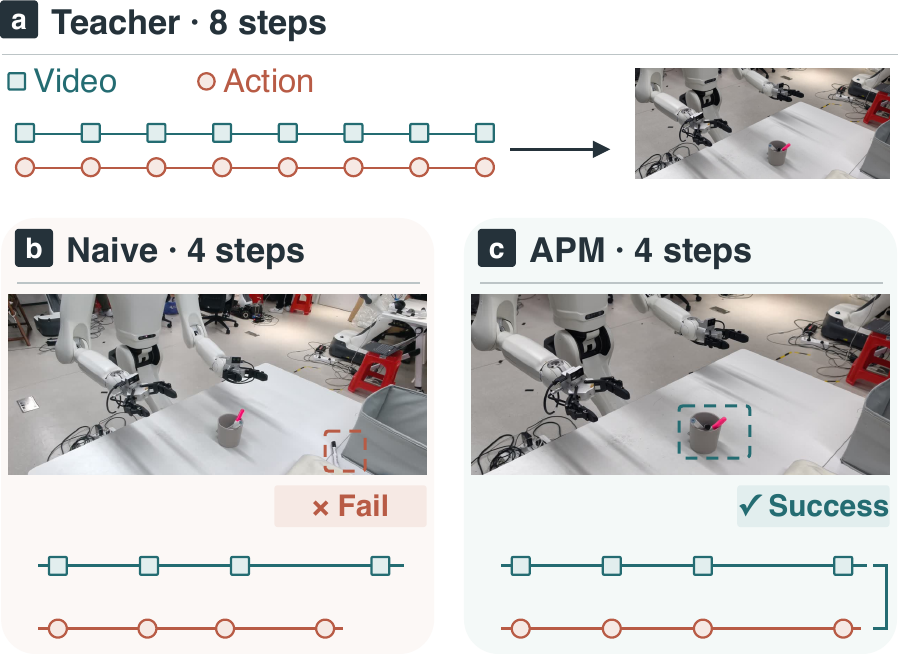}
\caption{\textbf{Student-state deviation in few-step generation.} Coarse student transitions can depart from the dense teacher trajectory and change the remaining prediction. APM constructs restart-anchor and local-field supervision at student-reached states. Naive and APM use the same initial condition and four steps; squares denote video states and circles denote action states.}
\label{fig:anchorstep_motivation}
\end{figure}

\section{Introduction}
\label{sec:introduction}

World Action Models (WAMs)~\cite{dreamzero,uwm,lingbot} jointly predict what a robot will see and do. Given the current observation, a language instruction, and the robot state, they extend vision-language-action policies~\cite{rt2,openvla} by generating future video and actions. However, iterative generation requires repeated model evaluations, leading to substantial inference costs. Few-step distillation addresses this limitation by training a student to approximate a multi-step teacher with fewer sampling steps~\cite{consistency_models,progressive_distillation,anyflow,flash_wam}. Supporting multiple sampling budgets with a single student is particularly desirable, as it allows the trade-off between inference efficiency and prediction quality to be adjusted at deployment without retraining.

Recent efforts improve few-step WAM distillation through teacher-guided transitions or endpoint supervision. AnyStep-WAM restarts the teacher from the first recycled preview state to supervise subsequent sampling-budget intervals, while its separate endpoint loss is computed from states along the teacher trajectory~\cite{anystep_wam}. RealtimeWAM incorporates teacher endpoint targets into action-consistency training under frozen video conditioning~\cite{realtimewam}. These approaches provide complementary forms of supervision, but their effectiveness across different sampling budgets remains an important challenge.

Specifically, coarse student updates can drive generation away from the teacher trajectory, causing the student to encounter states that the teacher would not normally visit. Consequently, local-transition supervision and endpoint supervision may be applied to different state distributions: a teacher transition provides guidance on local prediction dynamics, whereas an endpoint target constrains the completed video and actions. Supervision based on teacher-trajectory states may therefore be misaligned with the states encountered by the student, while local-transition guidance alone does not directly constrain the final output. Pairing these targets at the same student-reached state could address both limitations, but introduces an additional optimization challenge: their gradients may conflict when updating shared parameters, such that improving local transition matching can undermine endpoint predictions. Conventional weighted-loss objectives do not explicitly resolve this conflict, motivating a supervision framework that jointly aligns the two targets and prioritizes endpoint consistency.

To address these challenges, we propose \textbf{Anchored Path Matching (APM)}, which pairs endpoint and local-transition supervision at each student-reached state (Figure~\ref{fig:anchorstep_motivation}). A single teacher continuation from this state provides both targets: its endpoint anchors the completed video and actions, while its first intermediate state guides the local transition. Both student queries thus share the same starting state, without requiring an additional teacher rollout. To coordinate their optimization, we introduce {anchor-priority correction}, which preserves the endpoint gradient while removing conflicting components of the local-transition gradient within each modality. We adopt a two-stage training strategy, first distilling LingBot-VA~\cite{lingbot} using AnyFlow~\cite{anyflow}, and then freezing the resulting backbone while training modality-routed adapters with APM and retaining the base distillation objective. The resulting single checkpoint supports joint video--action generation with 1, 2, or 4 sampling steps, without teacher access at inference. 

On RoboTwin 2.0~\cite{robotwin2}, APM achieves average success rates of 80.9\% and 91.4\% at 1 and 4 steps, respectively, outperforming continued base distillation by 1.7 and 1.6 percentage points under matched trainable parameter counts and training budgets (Table~\ref{tab:main_results}). At one step, video PSNR improves from 30.169 to 32.317 dB, with consistent gains in SSIM and LPIPS. We further evaluate the effects of anchor selection, local-transition supervision, and gradient coordination through ablations, including fixed-state prediction evaluations. On two real-robot tasks, single-step APM also outperforms the step-matched undistilled teacher, reducing control-cycle latency from 4323.6 to 203.1 ms relative to the full teacher, a $21.3\times$ speedup.

Our contributions are summarized as follows:
\begin{itemize}
\item We propose {APM}, which pairs endpoint and intermediate teacher targets from a single continuation to supervise video--action predictions at the same student-reached state (Section~\ref{sec:stage2}).
\item We introduce {anchor-priority correction}, which prioritizes endpoint supervision by removing conflicting local-transition gradient components while preserving compatible guidance within each modality (Section~\ref{sec:correction}).
\item We demonstrate that a single APM checkpoint improves closed-loop success over continued base distillation at 1, 2, and 4 steps on RoboTwin 2.0, and outperforms the step-matched undistilled teacher on real-robot tasks (Section~\ref{sec:experiments}).
\end{itemize}

\begin{figure*}[t]
    \centering
    \includegraphics[width=\textwidth]{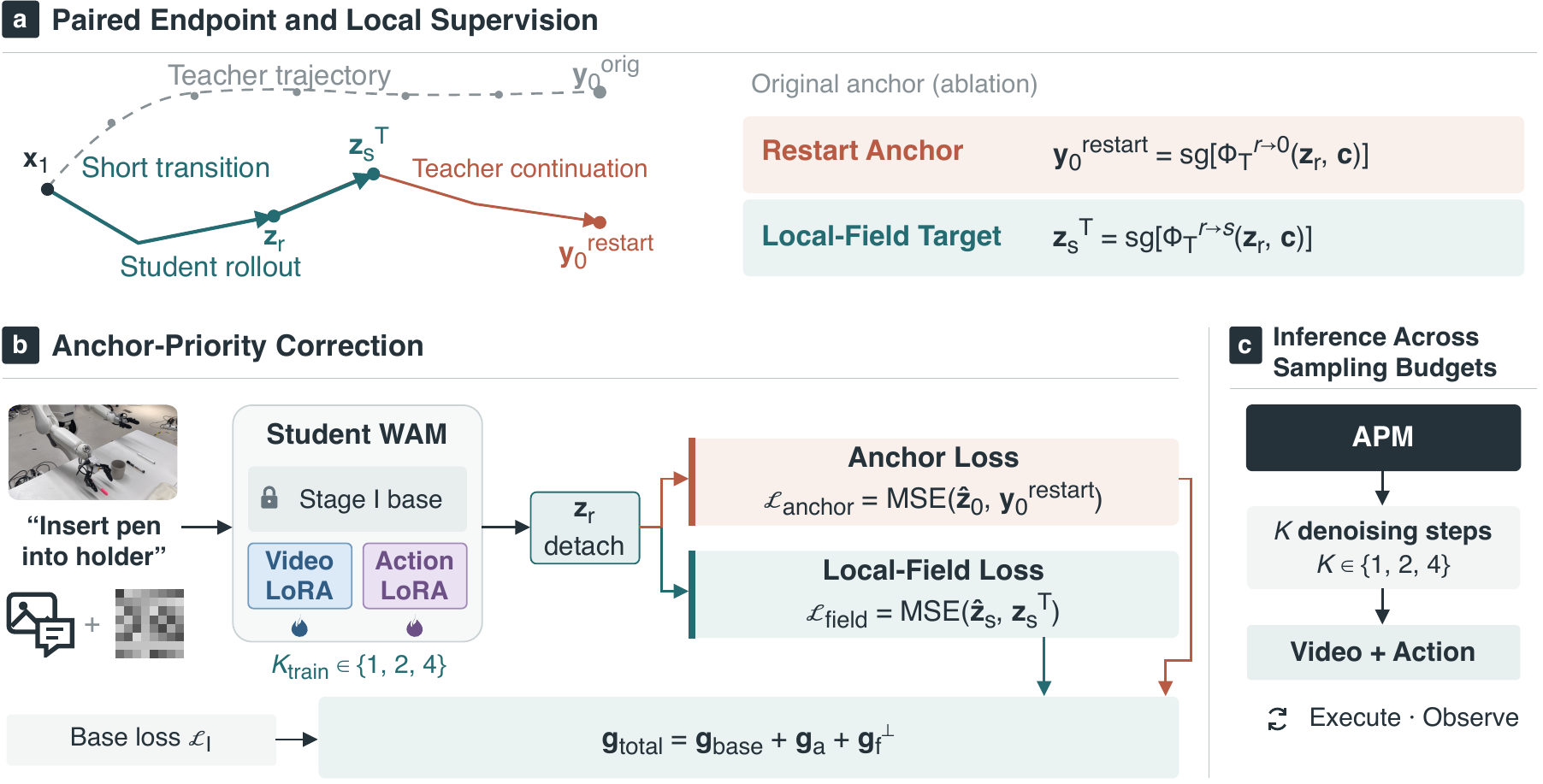}
    \caption{\textbf{Anchored Path Matching (APM).} (a) One teacher continuation from $\mathbf{z}_r$ supplies the restart endpoint and next nonzero-time local target. The gray path supplies the Original anchor used in ablation. $\Phi_T^{p\to q}$ integrates the teacher from noise level $p$ to $q$. (b) Endpoint and short-transition queries share $\mathbf{z}_r$; anchor-priority correction coordinates their gradients alongside the base objective. Image/text and noise icons denote $\mathbf{c}$ and $\mathbf{x}_1$; locks/flames denote frozen/trainable parameters. Modality superscripts are omitted. (c) The student generates video and action in $K$ steps and executes actions without teacher access.}
\label{fig:method_pipeline}
\end{figure*}

\section{Related Work}
\label{sec:related_work}

\paragraph{Efficient World Action Models.}
World Action Models (WAMs)~\cite{dreamzero,uwm,lingbot} jointly generate video and actions, extending vision-language-action policies~\cite{rt2,openvla,diffusion_policy,pi0} with visual prediction capabilities at additional inference cost. Existing approaches improve efficiency by removing video generation at deployment~\cite{fast_wam}, leveraging image-editing features~\cite{imagewam} or latent subgoals~\cite{lawam}, and adopting lightweight backbones~\cite{light_wam,efficient_wam}. Alternatively, few-step distillation reduces the number of denoising updates. Flash-WAM~\cite{flash_wam} adapts consistency learning to different video and action noise regimes, while AnyStep-WAM~\cite{anystep_wam} learns budget-aligned transitions and preview recovery for flexible inference. RealtimeWAM~\cite{realtimewam} incorporates teacher endpoint targets into action-consistency training. Unlike these approaches, APM pairs endpoint and finite-interval transition targets from a single teacher continuation at each student-reached state, with explicit gradient coordination.

\paragraph{Distillation and State Correction.}
Diffusion distillation accelerates sampling through progressive compression~\cite{progressive_distillation,tract}, trajectory rectification~\cite{rectified_flow,instaflow,perflow}, consistency learning~\cite{consistency_models,latent_consistency_models,ctm}, distribution matching~\cite{dmd,dmd2,sid}, and adversarial training~\cite{add}. Flow-map methods~\cite{flow_map_matching,align_your_flow} learn finite-interval transitions; MeanFlow~\cite{meanflow} parameterizes them through average velocity, while AnyFlow~\cite{anyflow} enables arbitrary-step video distillation. Shortcut Models~\cite{shortcut_models} further learn step-conditioned predictions without requiring a teacher. To address distribution shifts across successive predictions, Self Forcing~\cite{self_forcing} and Yang et al.'s WAM-OPD~\cite{wam_opd} train on student-generated execution histories. Within denoising trajectories, AnyFlow~\cite{anyflow} uses on-policy distribution matching, while Liu et al.'s WAM-OPD~\cite{wam_opd_pwtr} generates fresh denoising paths under replayed histories and applies execution-aware velocity supervision. Other approaches include teacher velocity regression~\cite{danceopd}, trajectory prediction with distribution correction~\cite{flow_map_without_data}, and consistency-based or flow-map-based supervision~\cite{fastopd,flowmap_opd}. APM complements these methods by jointly supervising endpoint predictions and finite-interval transitions from the same student-reached denoising state.

\paragraph{Multi-Objective Optimization.}
Endpoint and local losses can produce conflicting gradients. Loss weighting changes the relative strength of objectives~\cite{gradnorm,uncertainty_weighting}; gradient methods modify their joint update, with PCGrad~\cite{pcgrad} removing conflicting components and CAGrad~\cite{cagrad} balancing average descent with local task improvement. APM prioritizes the endpoint because it supervises the final video and actions. A one-sided projection preserves its gradient and removes the opposing component of the local gradient within each modality.

\section{Method}
\label{sec:method}

APM first distills LingBot-VA into a finite-interval video--action transition map, then corrects predictions at student-reached states (Figure~\ref{fig:method_pipeline}). With the base frozen, modality-routed adapters learn endpoint and local-transition targets from one teacher continuation. Anchor-priority correction coordinates their gradients alongside the base objective. One checkpoint supports 1-, 2-, and 4-step generation without teacher access.

\subsection{Setup and Student Initialization}
\label{sec:problem}
\label{sec:stage1}

Given a context $\mathbf{c}$ containing an observation, a language instruction, and the robot state, a WAM generates joint video and action latents $\mathbf{x}=(\mathbf{x}^v,\mathbf{x}^a)$ from a noisy prior $\mathbf{x}_1$. We use vectorized latents and conditioning inputs; $\boldsymbol{\theta}$ collects the student parameters. Each modality evolves from noise level 1 to 0. A joint state at clocks $\mathbf{t}=(t_v,t_a)$ is $\mathbf{x}_{\mathbf{t}}=(\mathbf{x}_{t_v}^v,\mathbf{x}_{t_a}^a)$; $\Psi_{\boldsymbol{\theta}}(\mathbf{x}_{\mathbf{t}};\mathbf{t}\!\to\!\mathbf{r},\mathbf{c})$ is its finite-interval transition to clocks $\mathbf{r}=(r_v,r_a)$. The target levels specify whether one joint denoising call predicts an endpoint or an intermediate state. A generation trajectory lies within one WAM call; a closed-loop rollout spans repeated calls and interactions with the environment.

Stage~I starts from the original LingBot-VA checkpoint~\cite{lingbot} and learns a finite-interval transition map using an AnyFlow-based distillation formulation~\cite{anyflow}. For modality $m\in\{v,a\}$, we construct the input from a training latent $\mathbf{x}_0^m$ and noise $\boldsymbol{\eta}^m$ as $\mathbf{x}_{t_m}^m=(1-t_m)\mathbf{x}_0^m+t_m\boldsymbol{\eta}^m$. The teacher and student predict from this common starting state.

\paragraph{Finite-interval queries.}
For source and target clocks $\mathbf{t}$ and $\mathbf{r}$ with $0\le r_m<t_m\le1$, the student directly predicts the state at the target clocks:
\begin{equation}
\hat{\mathbf{x}}_{\mathbf{r}}
=\Psi_{\boldsymbol{\theta}}(\mathbf{x}_{\mathbf{t}};\mathbf{t}\!\to\!\mathbf{r},\mathbf{c}).
\label{eq:transition-map}
\end{equation}
The target clocks select a completed output at zero noise or an intermediate state at nonzero noise.

\paragraph{Teacher-integrated targets.}
From the same input $\mathbf{x}_{\mathbf{t}}$, we numerically integrate the frozen teacher's predicted velocity field to $\mathbf{r}$. Let $\Phi_T^{\mathbf{t}\to\mathbf{r}}$ denote this integration. The detached target is
\begin{equation}
\mathbf{x}_{\mathbf{r}}^T
=\operatorname{sg}\!\left[
\Phi_T^{\mathbf{t}\to\mathbf{r}}(\mathbf{x}_{\mathbf{t}},\mathbf{c})\right],
\label{eq:stage1-target}
\end{equation}
where $\operatorname{sg}$ denotes stop-gradient and the superscript $T$ identifies a teacher target. The teacher uses eight integration steps to construct each target. Teacher-predicted velocities advance the state at each step. The student directly matches the integrated target using mean squared error (MSE):
\begin{equation}
\begin{aligned}
\mathcal L_m&=\operatorname{MSE}(\hat{\mathbf{x}}_{\mathbf{r}}^m,\mathbf{x}_{\mathbf{r}}^{T,m}),\\
\mathcal L_{\mathrm I}&=\mathcal L_v+\lambda_a\mathcal L_a.
\end{aligned}
\label{eq:stage1-loss}
\end{equation}
Here the superscript $m$ selects the modality component, and $\lambda_a$ balances the action term. Gradients pass through the student prediction, with the teacher target held fixed. Stage~II trains this initialized map to predict both endpoints and short transitions from states sampled along its own generation trajectories.

\subsection{Paired Endpoint and Local Supervision}
\label{sec:stage2}

\paragraph{Student-reached states.}
To align supervision with the states encountered during coarse generation, we construct both correction targets from the student's own rollout.
We select a correction state from a Stage~II rollout with budget $K_{\mathrm{train}}\in\{1,2,4\}$, writing $\mathbf{z}$ for these states (vs.\ $\mathbf{x}$ above). The one-step branch uses the initial state; the two- and four-step branches use intermediate states. Video and action use matching grids during correction, so a scalar $r$ abbreviates the joint clock $(r,r)$. Let $\mathcal R_{\boldsymbol{\theta},K_{\mathrm{train}}}^{1\to r}$ denote the rollout prefix. We detach its final state:
\begin{equation}
\mathbf{z}_r=(\mathbf{z}_r^v,\mathbf{z}_r^a)=\operatorname{sg}\!\left[
\mathcal R_{\boldsymbol{\theta},K_{\mathrm{train}}}^{1\to r}(\mathbf{x}_1,\mathbf{c})
\right].
\label{eq:student-reached-state}
\end{equation}
All correction queries start from this fixed state. Gradients update the student's predictions from $\mathbf{z}_r$ without propagating through the prefix that produced it.

\paragraph{Restart endpoint anchors.}
We continue the teacher from $\mathbf{z}_r$ to zero noise to obtain the endpoint target. For matched modality clocks, write $\Phi_T^{p\to q}$ for teacher integration from noise level $p$ to $q$ along the $K_T$-step grid, and $\hat{\mathbf{z}}_0^m=\Psi_{\boldsymbol{\theta}}^m(\mathbf{z}_r;r\!\to\!0,\mathbf{c})$ for the student's endpoint query. The restart anchor and its loss are
\begin{align}
\mathbf{y}_0^{\mathrm{restart}}&=\operatorname{sg}\!\left[
\Phi_T^{r\to0}(\mathbf{z}_r,\mathbf{c})\right],\label{eq:anchor-restart}\\
\mathcal L_{\mathrm{anchor}}^m
&=\operatorname{MSE}\!\left(\hat{\mathbf{z}}_0^m,\mathbf{y}_0^m\right).
\label{eq:loss-anchor}
\end{align}
Here $\mathbf{y}_0^m$ is the anchor's modality-$m$ component. The student queries $r\!\to\!0$ directly, while the teacher integrates the remaining steps without resampling noise. The gray path in Figure~\ref{fig:method_pipeline} is the Original anchor (Eq.~\ref{eq:anchor-orig}), used only in the ablation.

\paragraph{Local-transition guidance.}
The endpoint anchor constrains the completed output, while a short transition provides guidance closer to the starting state. We obtain this auxiliary target from the first intermediate state of the same teacher continuation. On the uniform teacher grid, for $s=r-1/K_T>0$ and $\hat{\mathbf{z}}_s^m=\Psi_{\boldsymbol{\theta}}^m(\mathbf{z}_r;r\!\to\!s,\mathbf{c})$, the local target and loss are
\begin{align}
\mathbf{z}_s^T&=\operatorname{sg}[\Phi_T^{r\to s}(\mathbf{z}_r,\mathbf{c})],\label{eq:field-target}\\
\mathcal L_{\mathrm{field}}^m
&=\operatorname{MSE}\!\left(\hat{\mathbf{z}}_s^m,\mathbf{z}_s^{T,m}\right).
\label{eq:loss-field}
\end{align}
The two losses supervise different intervals of the same transition map from a shared starting state. Caching the first teacher step supplies the local target without an additional rollout or a separate local velocity head.

For example, from the two-step student state $\mathbf{z}_{0.5}$, the teacher takes four remaining steps to zero noise. Its first step supplies the local target at $s=0.375$, and its endpoint supplies the anchor; the student learns each prediction through a separate query from $\mathbf{z}_{0.5}$.

\subsection{Anchor-Priority Correction}
\label{sec:correction}

\paragraph{Training objective.}
Stage~II freezes the base and trains modality-routed low-rank adaptation (LoRA) adapters in the video and action parameter groups. The frozen Stage~I base retains cross-modal context. Both modalities receive anchor and local-field supervision, with weighted correction losses
\begin{equation}
\mathcal L_{\mathrm{corr}}=
\sum_{m\in\{v,a\}}\left(
\lambda_{\mathrm{anchor}}^m\mathcal L_{\mathrm{anchor}}^m+
\lambda_{\mathrm{field}}^m\mathcal L_{\mathrm{field}}^m\right).
\label{eq:loss-correction}
\end{equation}
We retain the base objective $\mathcal L_{\mathrm I}$ and detach all teacher targets. We compute the weighted gradients separately and combine them using the update below. Section~\ref{sec:setup} gives the adapter ranks, weights, and sampling schedule.

\paragraph{Anchor-priority updates.}
\label{sec:optimization}
Although paired targets share an input state, their losses can produce conflicting gradients in shared parameters. We prioritize the anchor gradient because it supervises the completed output, removing the opposing component of the local-field gradient. Let $\mathbf{g}_{\mathrm{base}}$ be the gradient of $\mathcal L_{\mathrm I}$, and let $\mathbf{g}_a$ and $\mathbf{g}_f$ be the anchor and field gradients, each scaled by its loss weight. Here $a/f$ label objectives; modality indices are omitted. Within each modality parameter group, taking the inner product over that group, for $\mathbf{g}_a\ne\mathbf{0}$ we use
\begin{equation}
\begin{aligned}
\mathbf{g}_f^{\perp}
&=\mathbf{g}_f-\frac{\min(\langle \mathbf{g}_f,\mathbf{g}_a\rangle,0)}{\|\mathbf{g}_a\|_2^2}\mathbf{g}_a,\\
\mathbf{g}_{\mathrm{total}}&=\mathbf{g}_{\mathrm{base}}+\mathbf{g}_a+\mathbf{g}_f^{\perp}.
\end{aligned}
\label{eq:gradient-projection}
\end{equation}
If $\mathbf{g}_a=\mathbf{0}$, we set $\mathbf{g}_f^{\perp}=\mathbf{g}_f$. The superscript denotes conflict correction; orthogonality holds only when a conflicting component is removed. The projection acts on the field gradient; we then add the base gradient and apply correction at every optimization step.

\subsection{Inference Across Sampling Budgets}
\label{sec:inference}

At deployment, we choose $K\in\{1,2,4\}$ and use a matched video--action grid $t_i=1-i/K$, with joint clock $\mathbf{t}_i=(t_i,t_i)$. Starting from the prior $\mathbf{x}_{\mathbf{t}_0}=\mathbf{x}_1$ and current context $\mathbf{c}$, generation applies the learned transition map recursively:
\begin{equation}
 \mathbf{x}_{\mathbf{t}_{i+1}}=\Psi_{\boldsymbol{\theta}}\!\left(
 \mathbf{x}_{\mathbf{t}_i};\mathbf{t}_i\!\to\!\mathbf{t}_{i+1},\mathbf{c}\right),
 \quad i=0,\ldots,K-1.
\label{eq:inference-rollout}
\end{equation}
The endpoint $\mathbf{x}_{\mathbf{t}_K}=\mathbf{x}_0=(\mathbf{x}_0^v,\mathbf{x}_0^a)$ provides the predicted video and action latents. Equation~\ref{eq:inference-rollout} uses the same student weights at every budget; changing $K$ changes only the queried intervals and number of denoising calls. The robot executes an action chunk, refreshes its observation, and repeats, with $K$ fixed within each run. Stage~II trains all three budgets jointly, so choosing $K$ at deployment requires no retraining. Teacher integration and gradient projection are required only during training. Network function evaluations also count the classifier-free guidance (CFG) branches.

\section{Experiments}
\label{sec:experiments}


\subsection{Experimental Setup}
\label{sec:setup}

\paragraph{Tasks and Data.} We evaluate language-conditioned manipulation tasks on RoboTwin 2.0~\cite{robotwin2}, covering object insertion, placement, and assembly. Closed-loop performance is measured by task success rate under two test configurations: Clean and Random, where the latter applies domain randomization. All models are trained using Clean data and evaluated directly under Random conditions, without domain-randomized training data or test-time adaptation. For locally evaluated methods, we use identical tasks, evaluation seeds, episode counts, and inference budgets within each setting. Specifically, we evaluate 50 tasks with 50 episodes per task in each condition, resulting in 2,500 episodes per condition. Each episode uses a fixed environment initialization seed, giving 50 evaluation seeds per task and condition.

\paragraph{Training.} Stage~I is trained for 4,000 optimization steps using AnyFlow-based flow-map distillation~\cite{anyflow}, with $K_T=8$ teacher steps and $K_{\mathrm{train}}\in\{2,4\}$ student steps. We evaluate the Stage~II checkpoint after 5,000 additional optimization steps, giving 9,000 steps in total. In Stage~II, we freeze the Stage~I backbone and introduce modality-routed adapters, with an FFN rank of 64 and a temporal rank of 128. The batch size is 32.

\paragraph{Correction schedule.}
At each optimization step, we uniformly sample a correction budget from $K_{\mathrm{train}}\in\{1,2,4\}$. The one-step branch uses the initial state, while the two-step branch uses its sole intermediate state. For the four-step branch, we draw $b\sim\operatorname{Beta}(5,2)$ and select the state index $i$ and grid time $r$ as
\[
i=\min(1+\lfloor 3b\rfloor,3),\qquad r=1-i/4.
\]
This rule favors later states with lower noise.

Starting from the selected state, the teacher follows the remaining steps of its eight-step Euler grid. We use the resulting endpoint as the Restart anchor and cache the first teacher transition as the local target. Both correction losses use weights of 1.0 for video and 0.1 for action.

\paragraph{Inference.} We evaluate matched grids $K_{\mathrm{eval}} \in \{1, 2, 4\}$ (denoted $1/1$, $2/2$, $4/4$). Each student variant uses one checkpoint for all three budgets, matching the Stage~II correction settings. The LingBot-VA teacher and APM use the official LingBot-VA classifier-free guidance (CFG) parameters~\cite{lingbot}. The model-specific video evaluation settings are given in Section~\ref{sec:video_action}.

\paragraph{Baselines.} We compare APM against LingBot-VA~\cite{lingbot} evaluated with 1, 2, and 4 sampling steps to assess performance at matched inference budgets. We also include Flash-WAM~\cite{flash_wam}. Its single-step result is reported by DirectWAM~\cite{directwam}, while its two- and four-step results are obtained from our local evaluations. \textbf{Shared} denotes the Stage~I student before correction. \textbf{Continued} starts from the same Stage~I checkpoint and continues training with the base distillation objective only, using the same trainable parameter count, optimization settings, and total training steps as APM.

\begin{table}[!b]
\centering
\small
\setlength{\tabcolsep}{2pt}
\caption{\textbf{Manipulation performance on RoboTwin 2.0.} Clean uses standard conditions; Random applies domain randomization. Local evaluations use 50 tasks and 2,500 episodes per condition. Avg. averages the two success rates; $\Delta$ is the gain over Continued at the same budget (percentage points), computed before rounding. Bold identifies APM. V/A denotes video/action steps; each student uses one checkpoint. $\dagger$: reported by DirectWAM~\cite{directwam}; other Flash-WAM results are local.}
\label{tab:main_results}
\begin{tabular*}{\columnwidth}{@{\extracolsep{\fill}}lcrrrr@{}}
\toprule
Method & V/A & \multicolumn{3}{c}{Success rate (\%, $\uparrow$)} & $\Delta$ \\
\cmidrule(lr){3-5}
 & & Clean & Random & Avg. & \\
\midrule
LingBot-VA~\cite{lingbot} & $1/1$ & 57.0 & 50.7 & 53.9 & -- \\
Flash-WAM~\cite{flash_wam} & $1/1$ & 82.1$^{\dagger}$ & 74.2$^{\dagger}$ & 78.2$^{\dagger}$ & -- \\
Continued & $1/1$ & 82.3 & 76.1 & 79.2 & -- \\
\rowcolor{apmrow}\textbf{APM} & $\mathbf{1/1}$ & \textbf{84.3} & \textbf{77.5} & \textbf{80.9} & \textbf{+1.7} \\
\midrule
LingBot-VA & $2/2$ & 88.3 & 86.0 & 87.2 & -- \\
Flash-WAM & $2/2$ & 90.5 & 88.6 & 89.6 & -- \\
Continued & $2/2$ & 89.6 & 87.7 & 88.6 & -- \\
\rowcolor{apmrow}\textbf{APM} & $\mathbf{2/2}$ & \textbf{90.4} & \textbf{87.1} & \textbf{88.8} & \textbf{+0.1} \\
\midrule
LingBot-VA & $4/4$ & 90.4 & 90.0 & 90.2 & -- \\
Flash-WAM & $4/4$ & 92.0 & 90.4 & 91.2 & -- \\
Continued & $4/4$ & 90.5 & 89.1 & 89.8 & -- \\
\rowcolor{apmrow}\textbf{APM} & $\mathbf{4/4}$ & \textbf{92.5} & \textbf{90.3} & \textbf{91.4} & \textbf{+1.6} \\
\midrule
\multicolumn{6}{@{}l@{}}{\makebox[0pt][l]{Teacher~\cite{lingbot} ($25/50$, reported): Clean 92.9\%, Random 91.6\%.}} \\
\bottomrule
\end{tabular*}
\end{table}

\subsection{Manipulation Performance on RoboTwin 2.0}
\label{sec:main_results}

Table~\ref{tab:main_results} compares closed-loop manipulation performance across 1-, 2-, and 4-step video/action sampling budgets. Locally evaluated teacher and student models use matched evaluation conditions. The published single-step Flash-WAM result from DirectWAM~\cite{directwam} and the reported full-budget LingBot-VA results are included as reference points; the two- and four-step Flash-WAM results are obtained locally.

APM consistently improves average success over Continued across all three sampling budgets. At one step, APM achieves 80.9\% average success, a gain of 1.7 percentage points over Continued, with improvements of 2.0 and 1.4 points on Clean and Random, respectively. At two steps, APM achieves 88.8\% average success, a modest gain of 0.1 points; this corresponds to a 0.8-point improvement on Clean and a 0.6-point decrease on Random. At four steps, APM reaches 91.4\% average success, improving over Continued by 1.6 points, with gains of 2.0 points on Clean and 1.2 points on Random. Notably, all models are trained exclusively on Clean data. APM nevertheless improves Random success by 1.4 points at one step and 1.2 points at four steps, suggesting that the proposed correction also benefits generalization to domain-randomized conditions.

Each student variant uses the same weights across sampling budgets. The shared Stage~I student performs better with more sampling steps. APM improves its average success at each evaluated budget.

\begin{figure}[!htbp]
\centering
\includegraphics[width=\columnwidth]{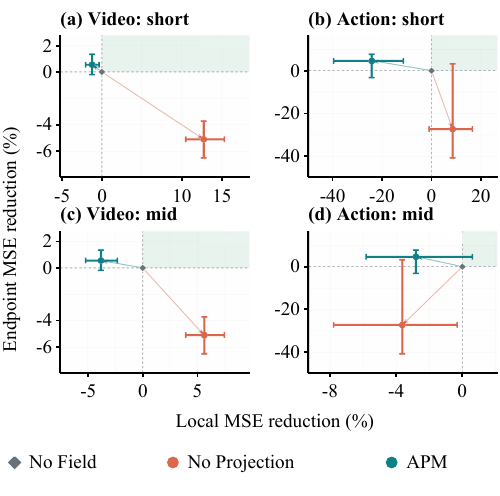}
\caption{\textbf{Local and endpoint prediction tradeoffs.} Reductions are relative to APM (Restart, No Field) at the origin. Rows: transitions $0.5\!\to\!0.375$ and $0.5\!\to\!0.25$, both paired with endpoint $0.5\!\to\!0$. Columns: video-latent and normalized-action MSE. Positive values indicate lower error; shading marks improvement on both axes. Bars show marginal 95\% paired bootstrap intervals, not joint regions. No Field omits local supervision; No Projection sums gradients; APM applies projection.}
\label{fig:fixed_state_tradeoff}
\end{figure}

\begin{figure*}[!t]
\centering
\includegraphics[width=0.96\textwidth]{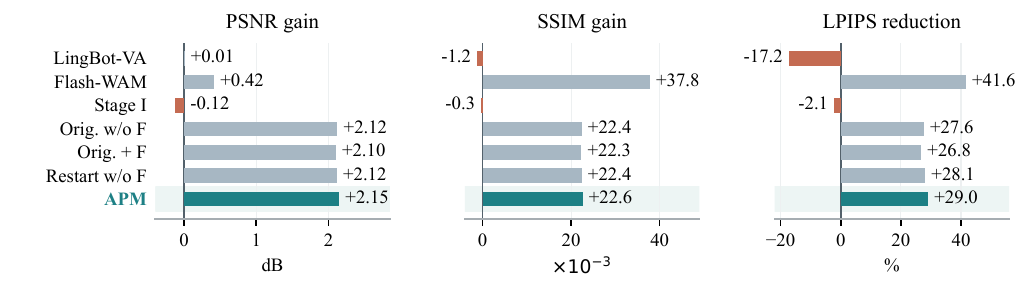}
\caption{\textbf{Single-step video quality gains over continued base distillation.} PSNR and SSIM show differences of means; LPIPS shows relative reduction of means. Positive values indicate improvement. Evaluation uses 500 shared contexts at $1/1$ video/action steps. Orig. denotes Original and F denotes field supervision; APM is Restart + Field.}
\label{fig:video_quality}
\end{figure*}

\subsection{Ablation Studies}
\label{sec:ablation}
\label{sec:fixed_state}

Table~\ref{tab:control_ablation} examines the effects of three design choices: endpoint anchor source, local-transition supervision (Field), and gradient projection (Proj.). All correction variants share the same Stage~I initialization and Stage~II training budget. We first compare Original and Restart anchors without local supervision, then introduce local supervision with projection for both anchor choices. Finally, with Restart and Field fixed, we compare direct gradient summation against anchor-priority projection.

\begin{table}[!htbp]
\centering
\small
\setlength{\tabcolsep}{2pt}
\caption{\textbf{Correction ablations.} Correction variants share initialization and training length. Field/Proj. enable local supervision/gradient projection. Cells report Clean/Random success (\%); columns give video/action steps. Shared and Continued denote initialization and continued distillation.}
\label{tab:control_ablation}
\begin{tabular*}{\columnwidth}{@{\extracolsep{\fill}}lccrrr@{}}
\toprule
Anchor & Field & Proj. & $1/1$ & $2/2$ & $4/4$ \\
\midrule
Shared & -- & -- & 82.6/75.5 & 89.5/87.3 & 90.6/90.0 \\
Continued & -- & -- & 82.3/76.1 & 89.6/87.7 & 90.5/89.1 \\
\midrule
Original & -- & -- & 84.2/76.2 & 90.4/87.9 & 90.5/89.8 \\
Original & $\checkmark$ & $\checkmark$ & 82.5/74.7 & 89.6/87.7 & 90.7/89.6 \\
\addlinespace[2pt]
Restart & -- & -- & 84.5/77.4 & 89.6/88.3 & 91.0/89.8 \\
Restart & $\checkmark$ & -- & 83.2/76.7 & 88.9/87.4 & 90.6/90.5 \\
\rowcolor{apmrow}\textbf{Restart} & $\boldsymbol{\checkmark}$ & $\boldsymbol{\checkmark}$ & \textbf{84.3/77.5} & \textbf{90.4/87.1} & \textbf{92.5/90.3} \\
\bottomrule
\end{tabular*}
\end{table}

The Original variant uses the teacher endpoint generated from the initial prior,
\begin{equation}
\mathbf{y}_0^{\mathrm{orig}}=\operatorname{sg}\!\left[
\Phi_T^{1\to0}(\mathbf{x}_1,\mathbf{c})\right],
\label{eq:anchor-orig}
\end{equation}
whereas Restart uses the endpoint constructed from $\mathbf{z}_r$ in Eq.~\ref{eq:anchor-restart}. Both variants share the same prior and context. When local supervision is enabled, they also use identical local-transition targets.

\textbf{Effect of the endpoint anchor source.}
We compare Original and Restart without local supervision to isolate the anchor effect. Restart improves Random success by 1.2, 0.4, and 0.0 percentage points at 1, 2, and 4 steps, respectively, while its effect on Clean success varies across budgets (+0.3, -0.8, and +0.5 points). This suggests that constructing the endpoint from the student-reached state is particularly beneficial under randomized conditions. 

\textbf{Effect of local-transition supervision.}
With Restart anchors, adding local supervision without projection changes Clean/Random success by (-1.3/-0.7), (-0.7/-0.9), and (-0.4/+0.7) percentage points at 1, 2, and 4 steps, respectively. Thus, local supervision alone does not consistently improve closed-loop performance, highlighting the need to coordinate it with endpoint supervision.

\textbf{Effect of anchor-priority projection.}
With Restart anchors and Field supervision fixed, we compare direct gradient summation (No Projection) against anchor-priority projection (APM). Projection improves average success by approximately 1.0, 0.6, and 0.9 percentage points at 1, 2, and 4 steps, respectively. This comparison isolates the benefit of prioritizing endpoint supervision when combining the two correction gradients.

\textbf{Fixed-state prediction analysis.}
Figure~\ref{fig:fixed_state_tradeoff} compares three Restart variants on identical states and teacher targets using 50 contexts with three fixed noise draws per context. We average predictions across noise draws and perform 10,000 paired task/episode bootstrap resamples with shared indices across models, modalities, and intervals. Projection reduces video endpoint error relative to unprojected Field supervision but increases local-transition error. Its endpoint improvement over No Field has a marginal 95\% percentile interval that crosses zero, confirming that improved local-transition accuracy does not necessarily translate into better endpoint predictions.

\subsection{Video Prediction Performance}
\label{sec:video_action}

We evaluate video generation independently of closed-loop control by conditioning each model on ground-truth observation histories at a $1/1$ video/action sampling budget. Figure~\ref{fig:video_quality} compares peak signal-to-noise ratio (PSNR), structural similarity (SSIM)~\cite{ssim}, and learned perceptual image patch similarity (LPIPS)~\cite{lpips}. The evaluation uses 500 contexts sampled from overlapping windows of 50 APM validation episodes, with one episode per RoboTwin 2.0 Clean task.

\textbf{Generation and scoring.}
We use the same Wan2.2 variational autoencoder (VAE) to separately decode the predicted and demonstration future latents, each appended to the corresponding ground-truth history. We evaluate the eight future frames by first averaging each metric over frames within a camera, then equally averaging across the three cameras and all contexts.

\begin{figure*}[!t]
\centering
\includegraphics[width=0.94\textwidth]{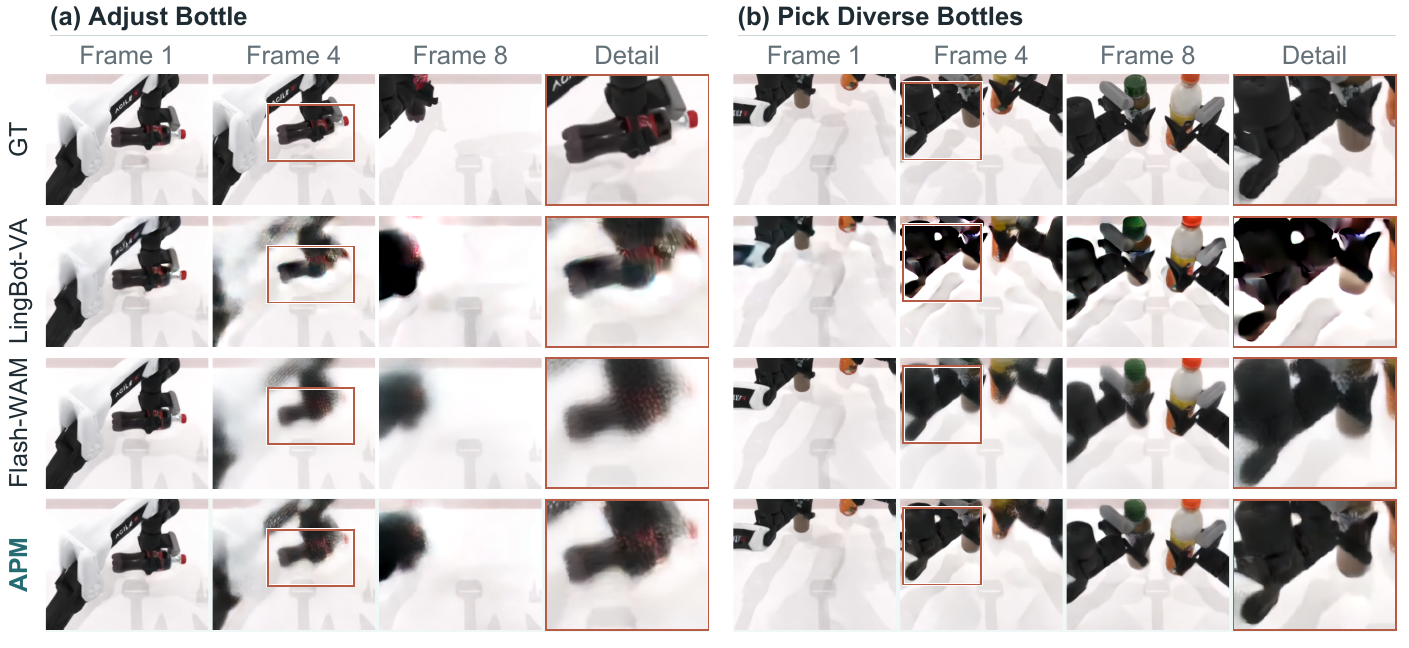}
\caption{\textbf{Qualitative comparison of single-step video predictions.} Rows show VAE-reconstructed ground truth (GT), LingBot-VA, Flash-WAM, and APM (Restart, Field); columns show head-camera frames 1, 4, and 8. Boxes mark frame-4 regions enlarged in the Detail column. From 500 contexts, we selected cases post hoc where all three head-camera metrics favor APM over Flash-WAM, then ranked them by LPIPS improvement and chose one per task.}
\label{fig:video_predictions}
\end{figure*}

\textbf{Model configurations.}
All APM ablations use the same Stage~II training budget. The video/action CFG scales are $5/1$ for LingBot-VA and APM, and $1/1$ for Flash-WAM, following the respective deployment configurations and the authors' guidance. Thus, each method is evaluated using its prescribed inference settings. For qualitative visualization, the eight predicted frames are displayed at 4 fps without interpolation.

\textbf{Quantitative results.} All APM variants improve PSNR, SSIM, and LPIPS over the locally evaluated LingBot-VA baseline. Compared with Continued, APM with Restart anchors and local-transition supervision increases single-step PSNR from 30.169 to 32.317 dB, with consistent improvements in SSIM and LPIPS. Compared with Flash-WAM under the respective model configurations, APM achieves higher PSNR, whereas Flash-WAM retains an advantage in SSIM and LPIPS.

\textbf{Effect of local supervision.}
The effect of local-transition supervision on video quality also depends on the endpoint anchor. With Restart anchors, adding local supervision and projection improves all three video metrics. In contrast, the same combination slightly degrades video quality with Original anchors. This is consistent with the closed-loop ablations and further supports pairing local supervision with an endpoint target constructed from the student-reached state. 
Figure~\ref{fig:video_predictions} shows qualitative predictions for Adjust Bottle and Pick Diverse Bottles. The enlarged regions highlight bottle and gripper contours in frame 4. In these selected examples, APM preserves clearer local structures than Flash-WAM.

\begin{figure}[!t]
\centering
\includegraphics[width=\columnwidth]{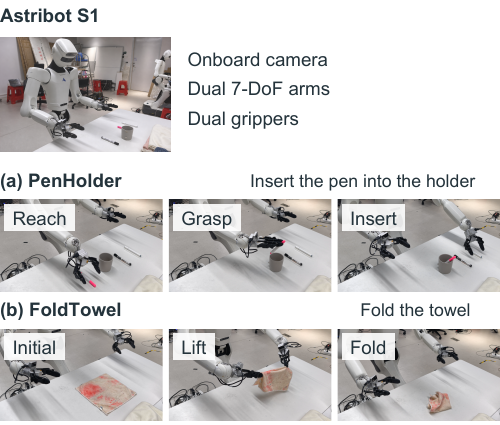}
\caption{\textbf{Real-robot platform and task execution.} Top: Astribot~S1 with an onboard camera, dual 7-DoF arms, and grippers. (a) \textsc{PenHolder}: reach, grasp, and insert. (b) \textsc{FoldTowel}: initial state, lift, and fold. Frames run from left to right.}
\label{fig:robot_platform}
\label{fig:real_robot_setup}
\end{figure}

\subsection{Real-Robot Deployment}
\label{sec:real_robot}

\begin{table}[!t]
\centering
\small
\setlength{\tabcolsep}{4pt}
\caption{Real-robot success rates and control-cycle latency. Each method uses 24 trials per task. V/A denotes video/action steps; bold identifies APM.}
\label{tab:real_robot}
\begin{tabular}{@{}lccc@{}}
\toprule
& \multicolumn{2}{c}{Success (\%, $\uparrow$)} & \\
\cmidrule(lr){2-3}
Method (V/A) & \textsc{PenHolder} & \textsc{FoldTowel} & \shortstack{Latency\\(ms, $\downarrow$)} \\
\midrule
Teacher (8/8) & 83.3 & 70.8 & 4323.6 \\
Coarse Teacher (1/1) & 29.2 & 25.0 & 187.5 \\
\textbf{APM (1/1)} & \textbf{62.5} & \textbf{75.0} & \textbf{203.1} \\
\bottomrule
\end{tabular}

\end{table}

\textbf{Robot platform.}
We conduct physical experiments using the Astribot~S1, equipped with an onboard camera, two 7-DoF arms, and dual grippers (Figure~\ref{fig:robot_platform}). Inference runs on an NVIDIA A800 GPU with 80~GB of memory. Without task-specific retraining, we compare the full teacher, the undistilled teacher evaluated on a coarse $1/1$ sampling grid, and APM at $1/1$. The evaluation covers two tasks, \textsc{PenHolder} and \textsc{FoldTowel}, with 24 trials per method for each task. Initial states are matched across methods.

\textbf{Quantitative results.}
Table~\ref{tab:real_robot} reports task success rates and control-cycle latency. Compared with the coarse-grid teacher, APM substantially improves success on both tasks, from 29.2\% to 62.5\% on \textsc{PenHolder} and from 25.0\% to 75.0\% on \textsc{FoldTowel}, with an additional 15.6 ms of latency. Compared with the full teacher, APM reduces control-cycle latency from 4323.6 to 203.1 ms, achieving a $21.3\times$ speedup. Its success rate is lower on \textsc{PenHolder} (62.5\% vs. 83.3\%) but higher on \textsc{FoldTowel} (75.0\% vs. 70.8\%). These results demonstrate that APM recovers substantial task performance under a one-step inference budget while retaining the latency advantage of coarse generation.

\textbf{Qualitative observations.}
Figure~\ref{fig:real_robot_setup} illustrates representative stages of physical task execution. Observed failure modes include grasp misalignment, cloth slippage, and insertion errors caused by pose drift. The coarse-grid teacher tends to fail during reaching and initial grasping, whereas APM failures occur more frequently in later manipulation stages.


\section{Conclusion and Discussion}
\label{sec:conclusion}

In this work, we introduced \textbf{Anchored Path Matching (APM)} for few-step World Action Model distillation. By pairing endpoint and local-transition supervision at the same student-reached state and prioritizing endpoint gradients, APM addresses both trajectory mismatch and conflicting supervision. A single checkpoint supports multiple sampling budgets without retraining or teacher access at inference. Experiments on RoboTwin 2.0 and physical robots demonstrate improved closed-loop manipulation and video prediction quality, together with substantially reduced inference latency while maintaining competitive task performance.

The current real-robot evaluation is limited to two manipulation tasks on a single platform. Both endpoint and local-transition targets are derived from the same teacher continuation, avoiding an additional teacher rollout for local supervision. Future work could adapt the correction according to teacher reliability and investigate longer prediction horizons and additional robot platforms.

{\small
\bibliographystyle{ieeenat_fullname}
\bibliography{references}
}
\end{document}